\documentclass[letterpaper]{article} 
\usepackage{aaai2026}  
\usepackage{times}  
\usepackage{helvet}  
\usepackage{courier}  
\usepackage[hyphens]{url}  
\usepackage{graphicx} 
\usepackage{natbib}  
\usepackage{caption} 
\usepackage{algorithm}
\usepackage{algorithmic}
\usepackage{multirow}
\usepackage{graphicx}
\usepackage{arydshln}
\usepackage[table]{xcolor}  
\usepackage{colortbl}
\usepackage{amsmath}

\usepackage{newfloat}
\usepackage{listings}
\DeclareCaptionStyle{ruled}{labelfont=normalfont,labelsep=colon,strut=off} 
\floatstyle{ruled}
\newfloat{listing}{tb}{lst}{}
\floatname{listing}{Listing}
\title{A Multi-Agent LLM Framework for Multi-Domain Low-Resource In-Context NER via Knowledge Retrieval, Disambiguation and Reflective Analysis}
\author{
    Wenxuan Mu\textsuperscript{\rm 1}, 
    Jinzhong Ning\textsuperscript{\rm 1}\footnotemark[1], 
    Di Zhao\textsuperscript{\rm 2}, 
    Yijia Zhang\textsuperscript{\rm 1}\thanks{Corresponding author.}\\
}
\affiliations{
    \textsuperscript{\rm 1}School of Information Science and Technology, Dalian Maritime University, Dalian, China\\
    \textsuperscript{\rm 2}School of Computer Science and Engineering, Dalian Minzu University, Dalian, China\\
    m1491377079@gmail.com, jinzhongning1996@gmail.com, zhaodi@dlnu.edu.cn, zhangyijia@dlmu.edu.cn
}
\usepackage{bibentry}

\makeatletter
\def\copyright@text{}
\makeatother

\begin{document}

\maketitle

\begingroup
\renewcommand\thefootnote{†}
\footnotetext{This paper has been accepted by AAAI 2026.}
\endgroup

\begin{abstract}
In-context learning (ICL) with large language models (LLMs) has emerged as a promising paradigm for named entity recognition (NER) in low-resource scenarios. However, existing ICL-based NER methods suffer from three key limitations: (1) reliance on dynamic retrieval of annotated examples, which is problematic when annotated data is scarce; (2) limited generalization to unseen domains due to the LLM's insufficient internal domain knowledge; and (3) failure to incorporate external knowledge or resolve entity ambiguities. To address these challenges, we propose KDR-Agent, a novel multi-agent framework for multi-domain low-resource in-context NER that integrates Knowledge retrieval, Disambiguation, and Reflective analysis. KDR-Agent leverages natural-language type definitions and a static set of entity-level contrastive demonstrations to reduce dependency on large annotated corpora. A central planner coordinates specialized agents to (i) retrieve factual knowledge from Wikipedia for domain-specific mentions, (ii) resolve ambiguous entities via contextualized reasoning, and (iii) reflect on and correct model predictions through structured self-assessment. Experiments across ten datasets from five domains demonstrate that KDR-Agent significantly outperforms existing zero-shot and few-shot ICL baselines across multiple LLM backbones. The code and data can be found at \url{https://github.com/MWXGOD/KDR-Agent}.
\end{abstract}


\section{Introduction}
\label{introduction}

Named entity recognition (NER) is a foundational task in information extraction \cite{NERsurvey}, supporting downstream applications such as relation extraction \cite{REsurvey}, question answering \cite{QAsurvey}, and knowledge graph construction \cite{QAsurvey}. Traditional NER methods typically depend on specialized neural architectures and extensive supervised fine-tuning on labeled datasets \cite{triNER1,triNER2,triNER3}. However, these approaches often exhibit limited generalization when transferred to new domains or entity types not seen during training \cite{triNERaffect}, presenting significant challenges in low-resource or emerging domain scenarios.

Recently, large language models (LLMs) \cite{GPT3,GPT4o,qwen,deepseekV3} have popularized in-context learning (ICL), which performs new tasks through prompts with few annotated demonstrations, eliminating parameter updates. Existing in-context NER approaches broadly fall into two lines. (1) \textbf{Few-shot ICL NER} assumes a manually labeled support set and, at inference time, retrieves a small number of relevant demonstrations from it to include in the prompt to perform NER
task \cite{GPTNER,codeie}. (2)\textbf{Zero-shot ICL NER} first prompts the LLM to label unlabeled texts to automatically build a provisional support set; at inference time, it likewise retrieves a few examples from this auto-labeled pool and typically applies filtering to reduce noise \cite{SINER,CMAS}.

Despite promising progress, in-context NER faces several limitations in multi-domain low-resource settings. (\textbf{\textit{Issue 1}}) Few-shot ICL NER assumes sufficient annotated examples are available for effective retrieval. In practice, particularly in low-resource settings, annotated data is scarce, reducing the effectiveness and increasing latency of the retrieval step. (\textbf{\textit{Issue 2}}) Zero-shot ICL NER relies heavily on the internal knowledge of LLMs regarding the target domain's entity type meaning. However, in novel or emerging domains, LLMs often lack sufficient background knowledge, hindering generalization. (\textbf{\textit{Issue 3}}) 
Both paradigms primarily focus on selecting demonstrations to include in prompts, but overlook the explicit need for external knowledge (e.g., biomedical terms or product names) and the resolution of ambiguous entity mentions (e.g., distinguishing ``Apple'' as a company or fruit). Consequently, mentions with limited local context frequently lead to classification errors. Effective performance in specialized domains thus requires external domain knowledge \cite{KECNER} and accurate disambiguation strategies \cite{context4NER}.

To overcome these challenges, we propose a novel \textbf{KDR-Agent} framework, a multi-agent LLM architecture explicitly designed for \emph{Knowledge Retrieval, Disambiguation, and Reflective Analysis} to enhance multi-domain, low-resource in-context NER.

For \textbf{Issue 1}, we observe that few-shot examples primarily help the model learn entity definitions and labeling guidelines (e.g., entity types and output formatting). Therefore, we propose explicitly providing concise, natural-language \emph{type definitions}, accompanied by a small, \textit{static set of few-shot demonstrations}. Unlike most few-shot ICL methods that rely solely on positive examples, we adopt an \emph{span-level positive–negative entity pairs contrastive} design to explicitly highlight common boundary and type confusions. This approach provides effective guidance while significantly reducing reliance on large annotated datasets, enabling effective low-resource NER across multiple domains.

To tackle \textbf{Issues 2 and 3}, KDR-Agent separates planning from execution. The \textit{central LLM planner} (i) determines which domain facts are missing and formulates targeted Wikipedia queries, and (ii) identifies entity mentions that are potentially ambiguous and require disambiguation. The \textit{Knowledge Retrieval Agent} then executes the queries and returns concise, source-attributed snippets from Wikipedia. Next, the \textit{Disambiguation Agent} resolves flagged ambiguities via brief self-check dialogues that leverage the local textual context. Finally, the \textit{Reflective Analysis Agent} conducts a structured review against predefined criteria and provides targeted feedback to refine the predictions. By explicitly injecting external domain knowledge and dedicating an agent to ambiguity resolution, KDR-Agent mitigates knowledge gaps and improves generalization in multi-domain, low-resource ICL NER.

In summary, the primary contributions of this paper are:
\begin{itemize}
    \item We propose KDR-Agent, a novel multi-agent LLM framework explicitly designed for multi-domain low-resource in-context NER tasks, effectively integrating external knowledge retrieval, entity mention disambiguation and reflective analysis for robust generalization.
    \item We introduce concise entity-type definitions combined with an entity-level positive-negative contrastive demonstration construction strategy, significantly reducing reliance on retrieving few-shot demonstrations from extensive labeled datasets.
    \item Experiments across multiple domains validate the effectiveness of KDR-Agent, demonstrating consistent improvements over state-of-the-art ICL-based NER methods.
\end{itemize}

\section{Related Work}

\subsection{In-Context NER}
Large language models (LLMs) exhibit strong semantic understanding and reasoning abilities \cite{GPT3,GPT4o,qwen,deepseekV3}. Through \textit{in-context learning} (ICL) \cite{iclsurvey}, they can perform tasks without parameter updates by following natural-language instructions, showing competitive zero-shot capabilities and improving further with a few prompt demonstrations. Recently, \textit{in-context NER} has attracted substantial attention. Early studies explored multi-turn prompt strategies with ChatGPT for zero-shot NER \cite{chatie,EZSNER}. Subsequent work found that inserting a small number of demonstrations into the prompt can significantly boost in-context NER performance. 

In most existing approaches, few-shot demonstrations are retrieved from a candidate support set using heuristics such as semantic similarity to the input. The support set can be constructed either from large collections of manually annotated examples \cite{GPTNER,codeie,C-ICL}, or by prompting an LLM to auto-label unlabeled texts and, at inference time, retrieving a few examples from this auto-labeled pool with additional filtering to reduce noise \cite{SINER,CMAS}.

Although C-ICL \cite{C-ICL} also explores contrastive in-context learning for information extraction with both positive and negative examples, our approach differs in how the demonstrations are constructed: we design entity-level positive–negative contrasts within the same instance, explicitly targeting boundary and type confusions. Empirical analyses are provided later.

\subsection{LLM-Based Multi-Agent IE Systems}

Recent advances in LLMs have spurred agent-based frameworks in which specialized agents collaborate via structured dialogue, tool use, and reflective feedback to solve complex reasoning tasks \cite{autogen,toolformer,Self-reflection,reflexion}. In information extraction (IE), initial explorations have adopted multi-agent paradigms to improve extraction quality. DAO \cite{DAO} employs a multi-agent optimization framework for event extraction, integrating external tools to enhance RAG quality and prediction reliability. CMAS \cite{CMAS} uses a cooperative multi-agent system to automatically label and filter unlabeled texts, yielding a candidate pool of few-shot demonstrations and improving zero-shot ICL NER.

In this paper, we focus on \textit{multi-domain, low-resource} in-context NER. The most related multi-agent NER work is CMAS \cite{CMAS}; however, it still faces the limitations highlighted in our Introduction: \textbf{Issue 2} (LLMs’ internal knowledge may not cover target domain types) and \textbf{Issue 3} (an emphasis on demonstration selection while overlooking explicit external knowledge requirements and systematic ambiguity resolution). Our KDR-Agent framework addresses these gaps by (i) planning targeted Wikipedia queries when domain knowledge is missing and (ii) dedicating agents to knowledge retrieval and ambiguity resolution, thereby improving generalization under low-resource, multi-domain conditions.

\begin{figure*}[t]
\centering
\includegraphics[width=2.1\columnwidth]{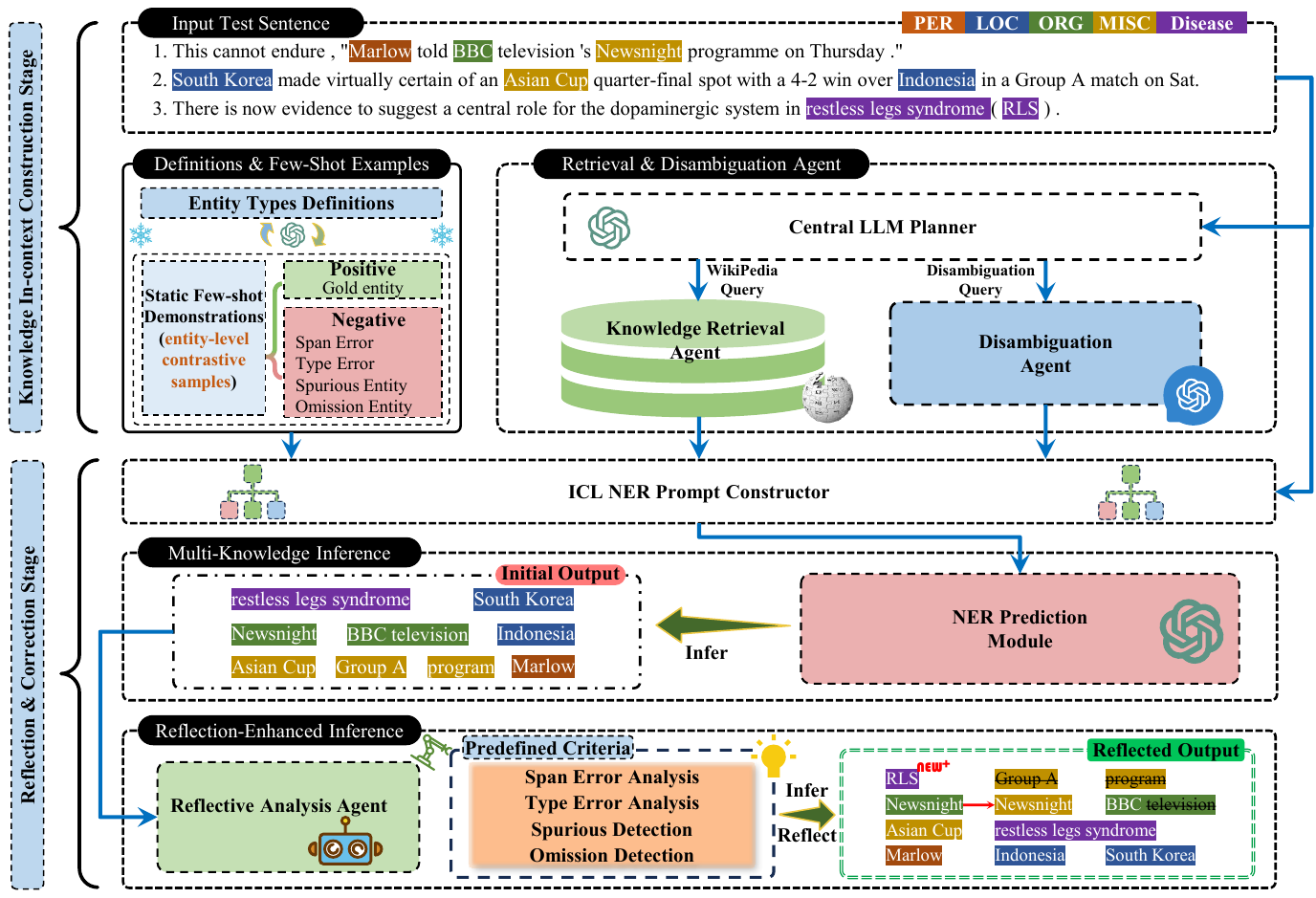} 
\caption{The overall architecture of the \textit{KDR-Agent} framework, consisting of two stages: (a) \textit{Knowledge In-context Construction} and (b) \textit{Reflection \& Correction}. The top illustrates the prompt construction process guided by external agents for \textit{knowledge retrieval} and \textit{disambiguation}, incorporating \textit{entity type definitions} and \textit{contrastive demonstrations}. The bottom shows the \textit{inference} and \textit{reflective correction} workflow based on predefined error criteria.}
\label{Overall_Model}
\end{figure*}

\section{Methodology}
\subsection{Task Formulation}

We study \textbf{multi-domain low-resource in-context NER}, where the model identifies entities from an input text $x = \{w_1, \dots, w_n\}$ under a unified type set $\mathcal{T}$. Unlike standard settings, only a small set of labeled examples $\mathcal{E} = {(x_j, y_j)}_{j=1}^{k}$ is available, shared across domains. The model performs NER via prompting LLMs, i.e., $\hat{y} = \text{LLM}(x \mid \mathcal{E})$, without fine-tuning. This setting requires the model to generalize across diverse domains ${\mathcal{D}_i}$ with limited supervision, often facing unfamiliar entity types and ambiguous mentions.

\subsection{Overall Framework of KDR-Agent}

We propose \textbf{KDR-Agent}, a multi-agent framework for enhancing in-context named entity recognition (NER) in low-resource, multi-domain settings. It integrates background knowledge, semantic disambiguation, and reflective correction into the prompt construction and inference process. As shown in Figure~\ref{Overall_Model}, KDR-Agent consists of two stages: (1) \textit{Knowledge In-context Construction}, which builds enriched prompts using type definitions, contrastive examples, and both retrieved background knowledge and disambiguation cues; and (2) \textit{Reflection \& Correction}, which performs structured error analysis to refine predictions. Specialized agents coordinate these steps to overcome key limitations of standard in-context NER.

\subsection{Stage 1: Knowledge In-context Construction}

\subsubsection{Natural-Language Type Definitions}

To reduce reliance on large domain-specific support sets, we introduce concise natural-language type definitions that guide the model’s understanding of entity categories. Let $\mathcal{T} = \{t_1, t_2, \dots, t_m\}$ denote the predefined entity types. For each $t_i \in \mathcal{T}$, we define a textual description $\mathcal{D}(t_i)$ capturing its semantic scope, including inclusion and exclusion criteria. We denote the concatenation of all type definitions as $\mathcal{P}_{\text{type}} = \texttt{Concat}(\mathcal{D}(t_1), \dots, \mathcal{D}(t_m))$, which serves as the \textit{type-definition section} in the final prompt.

These definitions are prepended to the prompt, enabling the model to align its predictions with intended type semantics, even in unfamiliar or low-resource domains. In practice, NER tasks are typically accompanied by annotation guidelines that define entity categories and labeling instructions; such definitions can be directly distilled into natural-language descriptions using an LLM, making this process broadly applicable and cost-effective.

\subsubsection{Static Few-Shot Contrastive Demonstrations}

We construct a static support set $\mathcal{E} = \{(x_j, y_j)\}_{j=1}^{k}$ to serve as few-shot in-context demonstrations. Each $x_j$ is a raw input sentence, and $y_j$ is the corresponding generation target, consisting of one or more entity mentions with their assigned types in a structured textual format (e.g., \texttt{``Barack Obama [PER]'', ``New York [LOC]''}).

To improve robustness, we introduce entity-level contrastive supervision by mixing correct and perturbed mention–type pairs. For each gold mention–type pair $(m_p, t_p)^+$ in $y_j$, we construct one contrastive negative pair $(m_n, t_n)^-$ by randomly applying one of four common error types: (1) mention boundary alteration (e.g., missing modifiers or partial names), (2) incorrect type label, (3) spurious mention that does not exist in $x_j$, or (4) omission of a valid mention from the output.
\begin{equation}
y_j = \{(m_p, t_p)^+\} \cup \{(m_n, t_n)^-\}
\end{equation}
where $(m_n, t_n)^-$ is constructed to simulate realistic annotation errors and guide the model to better distinguish type semantics and boundaries. These examples expose the model to boundary shifts, type confusions, hallucinated mentions, and omissions, promoting more accurate and calibrated generation.

We denote the formatted contrastive demonstrations from all $k$ examples as $\mathcal{P}_{\text{demo}} = \texttt{Format}(\mathcal{E})$, which serves as the \textit{few-shot demonstration section} in the final prompt. Each $(x_j, y_j)$ pair is serialized using a consistent output template  to ensure prompt consistency.

Unlike retrieval-based demonstration selection methods \cite{GPTNER,codeie}, this contrastive strategy enables in-context NER with only a small amount of manually annotated examples, eliminating the need for large labeled datasets.

\subsubsection{Central LLM Planner}

The Central LLM Planner serves as the controller of the multi-agent system, responsible for identifying knowledge gaps and coordinating downstream agents to improve prediction accuracy. Given an input text $x$ and the predefined entity type set $\mathcal{T} = \{t_1, t_2, \dots, t_m\}$, the planner performs two core reasoning steps.

First, it scans $x$ to detect concepts that may require external background knowledge for proper classification, such as domain-specific terminology or rare entity mentions. For each such concept $m_i$, the planner generates a search query $q_i = \texttt{GenerateQuery}(concept_i, x)$ and forwards it to the Knowledge Retrieval Agent.

Second, the planner identifies potentially ambiguous mentions—those that exhibit multiple plausible type assignments given limited context (e.g., ``Amazon'' as \textsc{ORG} or \textsc{LOC}). These mentions are flagged for disambiguation, and the planner constructs structured prompts to guide the Disambiguation Agent.

The planner produces two key outputs: (1) a query set $Q = \{ q_i \}$, containing the generated knowledge search queries; and (2) a disambiguation prompt segment $\mathcal{P}_{\text{ambig}}$, listing all flagged ambiguous mentions to be clarified by the Disambiguation Agent.

\subsubsection{Knowledge Retrieval Agent}

The Knowledge Retrieval Agent enriches the prompt with external factual knowledge to support entity disambiguation and classification. It takes as input the query set $Q = \{q_i\}$ generated by the Central LLM Planner and retrieves relevant information from Wikipedia.

For each query $q_i$, the agent issues a search request to Wikipedia and returns the corresponding introductory summary:
\begin{equation}
k_i = \texttt{RetrieveFromWiki}(q_i)
\end{equation}
where $k_i$ is the lead paragraph (introductory section) of the matched Wikipedia entry, providing concise and fact-based background for the queried concept. If no relevant page is found, the retrieval for $q_i$ fails and no knowledge snippet is returned.

The resulting knowledge set $\mathcal{K} = \{k_i\}$ is formatted into a structured prompt segment denoted as $\mathcal{P}_{\text{know}} = \texttt{Format}(\mathcal{K})$. In cases where all queries fail, $\mathcal{P}_{\text{know}}$ is left empty. When available, these retrieved summaries offer grounded external context to support the model's predictions on ambiguous or domain-specific mentions.

\subsubsection{Disambiguation Agent}

The Disambiguation Agent handles entity mentions $m_i$ that are flagged as ambiguous in $\mathcal{P}_{\text{ambig}}$, aiming to resolve their ambiguity through contextualized reasoning. It operates over each mention $m_i$ within the input $x$ and generates a natural language explanation that interprets the semantic role of $m_i$ in its specific context. This process is defined as:
\begin{equation}
\mathcal{P}_{\text{disamb}} = \texttt{Disambiguate}(x, \mathcal{P}_{\text{ambig}})
\end{equation}
where $\mathcal{P}_{\text{disamb}}$ is a set of \textit{concept–context interpretation statements}.

These explanations are inserted into the final prompt as contextual cues to guide the model in making more accurate type predictions during generation. If $\mathcal{P}_{\text{ambig}}$ is empty or no meaningful explanation can be produced, $\mathcal{P}_{\text{disamb}}$ is omitted.

\subsection{Stage 2: Reflection \& Correction}

In this stage, the model performs a two-step self-assessment and correction process. The goal is to improve prediction quality by identifying and revising potential errors in the initial output, guided by structured reflection.

\subsubsection{Initial Inference \& Multi-Knowledge Integration}

Given the input $x$ and the enriched prompt constructed in Stage 1, the LLM-based NER Prediction Module generates an initial output by integrating multiple sources of knowledge. The full prompt includes six components:
\begin{itemize}
\item  $x$: the target input text.

\item $\mathcal{P}_{\text{task}}$: task instruction describing the NER objective,  

\item $\mathcal{P}_{\text{type}}$: natural-language type definitions, 

\item $\mathcal{P}_{\text{demo}}$: contrastive few-shot demonstrations,  

\item $\mathcal{P}_{\text{know}}$: retrieved background knowledge,  

\item $\mathcal{P}_{\text{disamb}}$: disambiguation explanations,  
\end{itemize}

The model's initial generation is given by:
\begin{equation}
\hat{y}^{(0)} = \texttt{LLM}(x , \mathcal{P}_{\text{type}}, \mathcal{P}_{\text{demo}}, \mathcal{P}_{\text{know}}, \mathcal{P}_{\text{disamb}}, \mathcal{P}_{\text{task}})
\end{equation}

This prediction benefits from both external knowledge and structured in-context guidance, enabling context-aware entity extraction under low-resource conditions.

\subsubsection{Reflective Analysis Agent}

To identify and correct potential errors in the initial output $\hat{y}^{(0)}$, we introduce a \textit{Reflective Analysis Agent} that performs structured self-evaluation. This agent simulates post-hoc reasoning by comparing the generated predictions with the input $x$ and contextual signals from the prompt. Its goal is to detect common NER failure cases, justify them with linguistic or semantic evidence, and produce feedback for correction.

Specifically, we categorize common NER prediction errors into the following four types:

\textbf{Span Error}: The predicted mention is too broad or too narrow, resulting in boundary mismatch. For example, predicting “Barack” instead of “Barack Obama.”

\textbf{Type Error}: A correct mention is assigned an incorrect entity type. For instance, labeling “Apple” as a location rather than an organization.

\textbf{Spurious Detection}: The model predicts entities that are not mentioned or justified in the input, often due to hallucination or prior bias.

\textbf{Omission}: Valid entity mentions in the input are completely omitted from the prediction, typically due to subtle or implicit phrasing.

To perform this assessment, the agent is prompted with the input $x$, the model's initial output $\hat{y}^{(0)}$, and a reflection guideline $\mathcal{P}_{\text{Reflection}}$—a set of natural language instructions constructed based on the above four error types.

The reflection process is formalized as:
\begin{equation}
\mathcal{R} = \texttt{Reflect}(x, \hat{y}^{(0)}, \mathcal{P}_{\text{Reflection}})
\end{equation}
where $\mathcal{R}$ denotes a structured diagnostic report containing error labels, supporting justifications (e.g., span mismatch, missing mentions), and suggestions for revision.

The agent outputs $\mathcal{R}$ in natural language, which is then converted into a structured prompt segment $\mathcal{P}_{\text{reflect}}$ to support the final correction. This enables the model to revise its output through self-criticism, similar to how human annotators refine labels through multiple passes.

\subsubsection{Reflected Correction \& Final Output}

Based on the reflection report $\mathcal{R}$, we incorporate task-specific correction instructions $\mathcal{P}_{\text{Correction}}$ and perform a second round of prediction by prompting the LLM with an additional round of question–answer style reasoning, guided by the outputs of the Reflective Analysis Agent.

\begin{equation}
\hat{y}^{(1)} = \texttt{LLM}(x, \hat{y}^{(0)}, \mathcal{P}_{\text{Reflection}}, \mathcal{R}, \mathcal{P}_{\text{Correction}})
\end{equation}

The final prediction $\hat{y}^{(1)}$ integrates both multi-source knowledge and self-correction signals, enabling the model to revise earlier errors through explicit reasoning and guided reflection. As illustrated in Figure~\ref{Overall_Model}, this two-stage pipeline of initial generation followed by reflective correction produces more accurate and interpretable NER results, particularly under low-resource and multi-domain conditions.

\begin{table*}[t]
\centering
\resizebox{\textwidth}{!}{%
\begin{tabular}{clcccccccccc}
\hline
\multicolumn{2}{c}{} &
\multicolumn{2}{c}{\textbf{Biomedical}} &
\multicolumn{2}{c}{\textbf{Task-oriented Dialogue}} &
\multicolumn{2}{c}{\textbf{News}} &
\multicolumn{2}{c}{\textbf{Social Media}} &
\multicolumn{2}{c}{\textbf{Open-domain}} \\ \cline{3-12}
\multicolumn{2}{c}{\multirow{-2}{*}{\textbf{Model}}} &
\textbf{BC5CDR} & \textbf{NCBI} &
\textbf{MIT Movie} & \textbf{MIT Restaurant} &
\textbf{CoNLL-2003} & \textbf{OntoNotes 5.0} &
\textbf{Twitter Broad} & \textbf{Twitter NER-7} &
\textbf{WikiANN} & \textbf{WNUT-17} \\ \hline

\multicolumn{12}{c}{\cellcolor[HTML]{ECF4FF}\textit{\textbf{GPT-4o}}} \\ \hline

& ChatIE &
69.84 & 65.46 & 66.23 & 51.36 & 67.19 & 61.34 & 60.97 & 46.61 & 59.72 & 46.67 \\
& Self-Improving$^{\varDelta}$ &
72.16 & 70.13 & 66.74 & 50.16 & 74.68 & 61.60 & 65.01 & 48.21 & 58.16 & 48.21 \\
\multirow{-3}{*}{ZS} & CMAS$^{\varDelta}$ &
73.21 & 69.91 & 67.61 & 51.64 & 78.31 & 60.79 & 64.61 & 48.37 & 59.93 & 50.64 \\ \cdashline{1-12}

& GPT-NER$^{\varDelta}$ &
76.16 & 71.62 & 69.73 & 54.74 & 79.19 & 64.83 & 66.73 & 50.30 & 62.79 & 69.66 \\
& CodeIE$^{\varDelta}$ &
77.61 & 71.97 & 70.67 & 56.93 & 83.01 & 65.67 & 69.69 & 52.20 & 63.76 & 69.91 \\
\multirow{-3}{*}{FS} & \textbf{KDR-Agent}$^{\theta}$ &
\textbf{82.47} & \textbf{79.41} & \textbf{76.16} & \textbf{69.98} &
\textbf{83.34} & \textbf{71.85} & \textbf{74.90} & \textbf{60.87} &
\textbf{74.37} & \textbf{80.78} \\ \hline

\multicolumn{12}{c}{\cellcolor[HTML]{EFEFEF}\textit{\textbf{Qwen-2.5-72B}}} \\ \hline

& ChatIE &
66.91 & 62.12 & 62.31 & 47.68 & 60.85 & 58.42 & 58.33 & 43.20 & 56.71 & 44.15 \\
& Self-Improving$^{\varDelta}$ &
69.24 & 65.05 & 63.21 & 47.29 & 64.31 & 59.03 & 61.30 & 45.10 & 55.96 & 46.98 \\
\multirow{-3}{*}{ZS} & CMAS$^{\varDelta}$ &
70.15 & 65.61 & 62.60 & 48.12 & 63.12 & 58.88 & 61.75 & 45.76 & 57.22 & 48.02 \\ \cdashline{1-12}

& GPT-NER$^{\varDelta}$ &
73.24 & 68.81 & 66.40 & 52.84 & 66.58 & 62.20 & 64.01 & 47.82 & 60.41 & 66.55 \\
& CodeIE$^{\varDelta}$ &
76.10 & 72.95 & 67.91 & 56.61 & \textbf{73.85} & 64.32 & 66.10 & 52.21 & 63.93 & 70.26 \\
\multirow{-3}{*}{FS} & \textbf{KDR-Agent}$^{\theta}$ &
\textbf{81.45} & \textbf{79.22} & \textbf{76.16} & \textbf{67.70} &
73.73 & \textbf{67.37} & \textbf{73.55} & \textbf{59.99} &
\textbf{71.06} & \textbf{79.48} \\ \hline

\multicolumn{12}{c}{\cellcolor[HTML]{CBCEFB}\textit{\textbf{DeepSeek-V3}}} \\ \hline

& ChatIE &
65.34 & 60.44 & 62.51 & 45.91 & 62.77 & 56.32 & 56.41 & 41.86 & 54.92 & 42.81 \\
& Self-Improving$^{\varDelta}$ &
67.64 & 63.92 & 63.21 & 46.32 & 69.89 & 56.68 & 59.92 & 43.18 & 54.21 & 45.33 \\
\multirow{-3}{*}{ZS} & CMAS$^{\varDelta}$ &
68.12 & 63.27 & 63.99 & 46.90 & 72.71 & 56.71 & 59.62 & 44.27 & 55.62 & 46.13 \\ \cdashline{1-12}

& GPT-NER$^{\varDelta}$ &
72.40 & 67.81 & 66.78 & 54.71 & 74.98 & 60.77 & 62.52 & 46.91 & 59.03 & 65.14 \\
& CodeIE$^{\varDelta}$ &
73.45 & 68.23 & 67.34 & 58.11 & 75.44 & 61.56 & 64.87 & 48.10 & 60.65 & 65.81 \\
\multirow{-3}{*}{FS} & \textbf{KDR-Agent}$^{\theta}$ &
\textbf{81.38} & \textbf{78.66} & \textbf{76.31} & \textbf{65.99} &
\textbf{75.67} & \textbf{66.07} & \textbf{74.95} & \textbf{60.73} &
\textbf{70.44} & \textbf{78.52} \\ \hline

\end{tabular}%
}
\caption{Comparison of KDR-Agent and baseline methods across three LLM backbones (GPT-4o, Qwen-2.5-72B, and DeepSeek-V3) and five domains, covering ten benchmark NER test sets. \textit{ZS} and \textit{FS} denote Zero-shot and Few-shot ICL NER methods, respectively, as described in Section~\ref{introduction}. Superscript $^{\varDelta}$ indicates methods that construct prompts via retrieval-based demonstration selection, while $^{\theta}$ denotes the static few-shot demonstration strategy proposed in this paper.}
\label{mainresults}
\end{table*}

\section{Experiments and Analysis}
\subsection{Experiments Setting}
We evaluate the proposed \textbf{KDR-Agent} framework on ten benchmark NER datasets spanning five \textbf{\textit{Domains}}: \textbf{\textit{Biomedical}}—BC5CDR and NCBI \citep{BC5CDR,NCBI}; \textbf{\textit{Task-oriented Dialogue}}—MIT Movie and MIT Restaurant \citep{MITMoviesNER,Mitrestuarant}; \textbf{\textit{News}}—CoNLL-2003 and OntoNotes 5.0 \citep{conll2003,ontonotes4}; \textbf{\textit{Social Media}}—Twitter Broad and Twitter NER-7 \citep{BroadTwitter,TweetNER7}; and \textbf{\textit{Open-domain}}—the English subset of WikiANN and WNUT-17 \citep{WikiANN,WNUT‑17}.  This diverse collection enables a comprehensive evaluation of the framework’s generalization ability across a wide range of specialized domains under low-resource conditions. We report the F1-score as the primary evaluation metric to assess entity recognition performance.

For all datasets, we use the development sets solely for hyperparameter tuning and evaluate performance on the official test sets. An exception is WikiANN, where we randomly sample 5,000 instances from both the development and test sets due to its large scale. For the other datasets, the full development and test sets are used. We set the number of few-shot demonstrations to 10 for the datasets with a larger number of entity types, including MIT Movie NER, MIT Restaurant NER, and OntoNotes 5.0, and to 5 for the remaining datasets. In the Central LLM planner, we limit the number of Wikipedia query terms and mentions to be disambiguated to a maximum of 5. For reproducibility, we employ the MediaWiki Action API to retrieve knowledge from Wikipedia, restricting all queries to information available before May 1, 2025, and keeping only the summary section of each retrieved article.

\subsection{LLM Backbones and Baselines}
In this paper, we evaluate the performance of KDR-Agent and several baseline methods on top of representative large language models (LLMs), including the proprietary GPT-4o \cite{GPT4o} and two open-source backbones: DeepSeek-V3 \cite{deepseek} and Qwen-2.5-72B \cite{qwen2}. We compare KDR-Agent with two categories of representative in-context NER baselines:

\subsubsection{Zero-shot ICL NER}
\begin{itemize}
    \item \textbf{ChatIE}~\cite{chatie}: Reformulates the zero-shot IE task as a multi-turn question answering problem using a two-stage chat-based LLM framework.
    \item \textbf{Self-Improving}~\cite{SINER}: Prompts the LLM to generate predictions on unlabeled data via self-consistency, and designs selection strategies to identify reliable demonstrations for ICL-based NER.
    \item \textbf{CMAS}~\cite{CMAS}: Employs a cooperative multi-agent system built on LLMs to automatically annotate unlabeled data and select filtered examples as demonstrations for zero-shot ICL NER.
\end{itemize}

\subsubsection{Few-shot ICL NER}
\begin{itemize}
    \item \textbf{GPT-NER}~\cite{GPTNER}: Retrieves annotated examples based on semantic similarity to the input for constructing few-shot ICL prompts. Unlike the original work, which also includes parameter tuning, we only use its prompting strategy for ICL-based inference.
    \item \textbf{Code-IE}~\cite{codeie}: Designs code-style prompts to structure both input and output formats, enabling the construction of structured few-shot demonstrations for in-context NER.
\end{itemize}

\subsection{Main Results}

Table~\ref{mainresults} presents the performance comparison of KDR-Agent against a series of in-context NER baselines across ten benchmark datasets spanning five domains, under three representative LLM backbones: GPT-4o, Qwen-2.5-72B, and DeepSeek-V3. Overall, KDR-Agent consistently outperforms both zero-shot (ZS) and few-shot (FS) baselines across all domains and model backbones. From the results, we have the following observations: 

(1) \textbf{KDR-Agent significantly outperforms all few-shot (FS) baselines}. Compared to methods like GPT-NER and Code-IE, which rely on retrieving demonstrations from large annotated corpora, KDR-Agent uses a static set of carefully designed positive-negative contrastive demonstrations. This eliminates the need for retrieval, reduces latency, and enhances robustness in low-resource scenarios—directly addressing \textit{Issue 1} in the introduction.

(2) \textbf{The overall performance of KDR-Agent and all few-shot (FS) baselines surpasses that of the zero-shot (ZS) baselines}, supporting our observation in the introduction that demonstrations help the LLM better understand task-specific labeling guidelines, such as entity type definitions and output formatting. This in-context learning of demonstration patterns enables few-shot methods like KDR-Agent to compensate for the internal knowledge limitations of zero-shot approaches—addressing \textit{Issue 2}.

(3) \textbf{KDR-Agent achieves especially notable improvements in complex domains} such as \textit{Biomedical} and \textit{Social Media}, where domain-specific knowledge and semantic disambiguation are crucial. By dynamically incorporating external knowledge and performing local context-aware disambiguation via its multi-agent collaborative framework, KDR-Agent enhances generalization across specialized NER scenarios—tackling \textit{Issue 3} outlined in the introduction.

\begin{table}[t]
\centering
\resizebox{\columnwidth}{!}{%
\begin{tabular}{lccc}
\hline
\textbf{}          & \multicolumn{1}{l}{\textbf{NCBI}} & \multicolumn{1}{l}{\textbf{OntoNotes 5.0}} & \multicolumn{1}{l}{\textbf{Twitter NER-7}} \\ \hline
\textbf{KDR-Agent} & \textbf{79.41}                    & \textbf{71.85}                             & \textbf{60.87}                             \\ \cdashline{1-4}
-Reflection        & 75.91                             & 70.17                                      & 57.81                                      \\
-KRA               & 76.21                             & 71.70                                      & 59.34                                      \\
-DA                & 75.49                             & 70.73                                      & 55.81                                      \\
-KRA\&DA            & 74.16                             & 69.94                                      & 55.07                                      \\
-NS                & 78.36                             & 70.69                                      & 58.99                                      \\ \hline
\end{tabular}%
}
\caption{
Ablation study results on the GPT-4o LLM backbone (F1 score). \textit{Reflection} denotes the Reflection \& Correction Stage in KDR-Agent, \textit{KRD} refers to the Knowledge Retrieval Agent, \textit{DA} indicates the Disambiguation Agent, and \textit{NS} represents the entity-level negative contrastive samples used in few-shot demonstrations.
}
\label{ablationstudy}
\end{table}

\begin{figure*}[t]
\centering
\includegraphics[width=\textwidth]{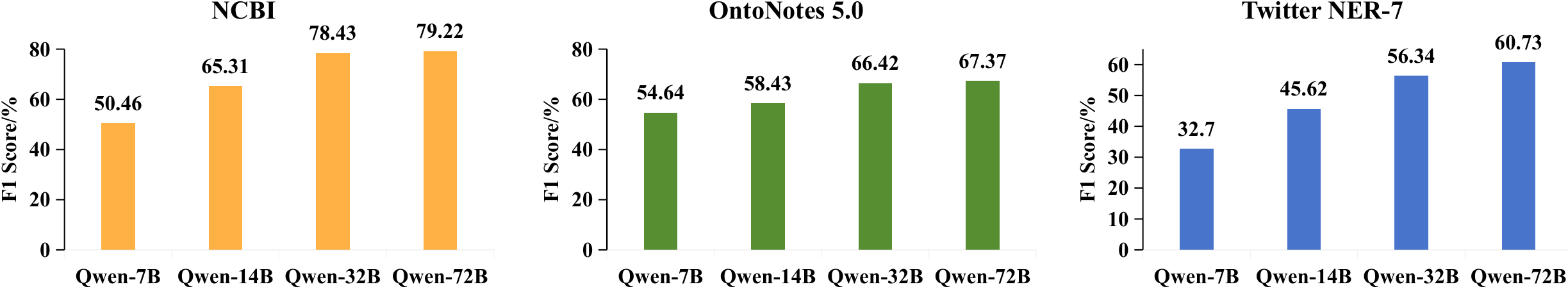} 
\caption{F1 performance of KDR-Agent on three datasets with different sizes of Qwen LLM backbones.}
\label{scaling_law}
\end{figure*}

\subsection{Ablation Study}

To evaluate the contribution of each component in \textbf{KDR-Agent}, we conduct ablation experiments on three representative datasets from distinct domains: NCBI (Biomedical), OntoNotes 5.0 (News), and Twitter NER-7 (Social Media), using GPT-4o as the backbone. The results are presented in Table~\ref{ablationstudy}.

Removing the Reflection \& Correction stage consistently results in performance drops across all domains, highlighting its importance in refining outputs through structured self-review. This stage helps the model handle borderline cases and enhances the overall stability of predictions.

When the Knowledge Retrieval Agent and/or the Disambiguation Agent are removed, we observe notable performance degradation on domain-specific datasets, especially in the biomedical and social media domains. These results indicate that \textbf{different domains benefit from different components}: biomedical NER relies heavily on both domain-specific knowledge and disambiguation, while social media NER is more dependent on effective ambiguity resolution. In contrast, general-purpose domains such as news show relatively minor changes, suggesting lower reliance on domain-specific knowledge or disambiguation, though still benefiting from the overall multi-agent design.

Additionally, removing the negative contrastive examples in few-shot demonstrations leads to modest but consistent performance declines. This demonstrates that \textbf{contrastive examples help the model better distinguish between confusing entity types and boundaries}, thereby improving robustness in low-resource scenarios.

\subsection{Impact of Backbone Scale on Performance}

We investigate the impact of reducing the LLM backbone size on in-context NER performance by evaluating KDR-Agent with four variants of the Qwen model at different parameter scales. As shown in Figure~\ref{scaling_law}, the overall performance shows a clear downward trend as the model size decreases. This indicates that smaller models exhibit reduced semantic understanding and reasoning capacity, which are critical for effective in-context learning.

The performance degradation is more pronounced on the biomedical dataset (NCBI) and the social media dataset (Twitter NER-7), while the drop on the news dataset (OntoNotes 5.0) is relatively moderate. This observation suggests that complex domains with specialized terminology or high ambiguity rely more heavily on the backbone model's reasoning ability and domain adaptability. In contrast, general-purpose domains such as news require less domain-specific inference and are therefore less sensitive to model scale.

These results indicate that while smaller LLMs offer computational efficiency, their performance may be inadequate for complex domains that demand strong semantic reasoning and disambiguation capabilities.

\begin{table}[t]
\centering
\resizebox{\columnwidth}{!}{%
\begin{tabular}{lccc}
\hline
\textit{}            & \multicolumn{1}{l}{\textbf{NCBI}} & \multicolumn{1}{l}{\textbf{OntoNotes 5.0}} & \multicolumn{1}{l}{\textbf{Twitter NER-7}} \\ \hline
\multicolumn{1}{c}{} & \multicolumn{3}{c}{\textit{Span Error Rate(\%)}}                                                                             \\ \hline
w/o Reflection       & 22.03                             & 17.33                                      & 9.86                                       \\
+ Reflection         & 9.18                              & 12.55                                      & 7.09                                       \\ \hline
\multicolumn{1}{c}{} & \multicolumn{3}{c}{\textit{Type Error Rate(\%)}}                                                                             \\ \hline
w/o Reflection       & -                                 & 6.14                                       & 17.22                                      \\
+ Reflection         & -                                 & 4.49                                       & 7.78                                      \\ \hline
\multicolumn{1}{c}{} & \multicolumn{3}{c}{\textit{Spurious Detection Rate(\%)}}                                                                     \\ \hline
w/o Reflection       & 16.44                             & 24.19                                      & 24.27                                      \\
+ Reflection         & 5.57                              & 17.67                                      & 12.57                                      \\ \hline
\multicolumn{1}{c}{} & \multicolumn{3}{c}{\textit{Omission Detection Rate(\%)}}                                                                     \\ \hline
w/o Reflection       & 49.62                             & 24.94                                      & 48.97                                      \\
+ Reflection         & 17.78                             & 20.64                                      & 30.38                                      \\ \hline
\end{tabular}%
}
\caption{
Error analysis of KDR-Agent on three datasets, with and without the Reflection \& Correction stage, using GPT-4o as the backbone. Each error rate is reported as the proportion of predicted entities. The four error types include span errors, type errors, spurious detections, and omissions. \textit{Note:} The NCBI dataset contains only a single entity type, and thus type errors are not applicable in this case.
}
\label{erroranalysis}
\end{table}

\subsection{Error Analysis}

We conduct an error analysis on three datasets to assess the impact of the Reflection \& Correction stage in KDR-Agent. Errors are categorized into \textit{span errors}, \textit{type errors}, \textit{spurious detections}, and \textit{omissions}, each measured as a proportion of predicted entities. As shown in Table~\ref{erroranalysis}, the reflection module consistently reduces all error types, with the most notable improvements in spurious detections, type errors, and omissions—particularly on biomedical and social media datasets. These gains highlight the module’s effectiveness in correcting type assignments and recovering missed entities in domains characterized by complex terminology or informal expressions. In contrast, improvements on the news domain are modest, likely due to its syntactic regularity and semantic clarity. Overall, the results demonstrate that the Reflection \& Correction stage substantially improves prediction quality, especially in domains where disambiguation and boundary precision are critical.


\section{Conclusion}

We propose KDR-Agent, a multi-agent LLM framework for low-resource in-context NER across diverse domains. By integrating external knowledge retrieval, ambiguity disambiguation, and reflective analysis, our method mitigates domain knowledge gaps and reduces reliance on dynamic example retrieval. Extensive experiments demonstrate that KDR-Agent consistently outperforms strong baselines, highlighting its effectiveness and generalizability in challenging NER settings.

\section{Acknowledgements}
This work was supported in part by the Young Scientists Fund of the National Natural Science Foundation of China (NSFC) under Grant No. 62506058 and the Fundamental Research Funds for the Central Universities – Young Teacher Scientific and Technological Innovation Project under Project No. 3132025277.

\bibliography{aaai2026}

\end{document}